\documentclass[sigconf]{acmart}

\usepackage{amsmath}
\usepackage{booktabs}
\usepackage{multirow}
\usepackage{placeins}
\usepackage{cuted}
\usepackage{algorithm}
\usepackage{algpseudocode}
\usepackage{xspace}

\newcommand{\kgn}{KG-Neighbor\xspace}
\newcommand{\idprofile}{Instance Diagnosis Profile\xspace}
\newcommand{\idp}{IDP\xspace}

\newcommand{\Demb}{D_{\mathrm{emb}}}
\newcommand{\Dkg}{D_{\mathrm{kg}}}
\newcommand{\Dkgn}{D_{\mathrm{KGN}}}
\newcommand{\Skg}{S_{\mathrm{kg}}}

\newcommand{\pir}{\ensuremath{\mathrm{PI}}\xspace}
\newcommand{\lir}{\ensuremath{\mathrm{LI}}\xspace}
\newcommand{\bcr}{\ensuremath{\mathrm{BC}}\xspace}
\newcommand{\ucr}{\ensuremath{\mathrm{UC}}\xspace}

\setcopyright{none}
\renewcommand\footnotetextcopyrightpermission[1]{}

\acmConference[WSDM '27]{The 20th ACM International Conference on Web Search and Data Mining}{February 15--19, 2027}{Hong Kong, China}

\begin{document}

\title{KG-Neighbor: Knowledge-Guided Neighborhood Construction for Embedding Visualization}

% Author information is hidden automatically in anonymous review mode.
\author{Chang Won Kim}
\affiliation{%
  \institution{DATUMO Inc.}
  \city{Seoul}
  \country{Republic of Korea}}
\email{changwon.kim@selectstar.ai}

\author{Jaewook Lee}
\affiliation{%
  \institution{DATUMO Inc.}
  \city{Seoul}
  \country{Republic of Korea}}
\email{jaewook.lee@selectstar.ai}

\author{Haeun Min}
\affiliation{%
  \institution{DATUMO Inc.}
  \city{Seoul}
  \country{Republic of Korea}}
\email{haeun.min@selectstar.ai}

\author{JunNyeong Jeong}
\affiliation{%
  \institution{DATUMO Inc.}
  \city{Seoul}
  \country{Republic of Korea}}
\email{junnyeong.jeong@selectstar.ai}

\author{Minwoo Sim}
\affiliation{%
  \institution{DATUMO Inc.}
  \city{Seoul}
  \country{Republic of Korea}}
\email{minwoo.sim@selectstar.ai}

\author{Barom Kang}
\affiliation{%
  \institution{DATUMO Inc.}
  \city{Seoul}
  \country{Republic of Korea}}
\email{rkdqkfha123@selectstar.ai}

\author{Kyohoon Jin}
\correspondingauthor
\affiliation{%
  \institution{DATUMO Inc.}
  \city{Seoul}
  \country{Republic of Korea}}
\email{kyohoon.jin@selectstar.ai}

\renewcommand{\shortauthors}{Kim et al.}

\begin{abstract}
In embedding visualizations, nearby instances are commonly interpreted as semantically related, although neighborhoods derived from input embeddings may not fully reflect relationships in instance content. Neighborhood-based dimensionality reduction (DR) methods can propagate this limitation because they optimize low-dimensional layouts from embedding-based neighborhood graphs. We propose Knowledge-Guided Neighborhood Graph Construction (\kgn), which represents relational knowledge derived from each instance as a set of knowledge graph (KG) triples. It uses the degree of triple sharing as pairwise semantic evidence to adjust embedding distances and reconstruct the top-$k$ neighborhood graph; neither KG extraction nor neighborhood construction uses class labels. We also introduce \idprofile{} (\idp), which interprets residual mismatches by independently assessing whether the source content supports the class label provided by the data set, referred to as the assigned class, and the neighborhood-majority class. Across eight text and image dataset--encoder conditions, we characterize \kgn over the full guidance-strength sweep and four KG similarity measures. Across this sweep, the maximum NMI and $\mathrm{LT}_{2D}$ each exceed the corresponding kNN baseline in all eight conditions, with the largest observed gains reaching 0.6871 and 1.3563, respectively. The gains and the extent of neighborhood reconfiguration vary across conditions and are larger under encoder conditions with lower kNN baseline scores. The \idp analysis shows that the largest reductions are concentrated in Placement Issue candidates, whereas Label Issue and Unresolved Case instances account for a larger share of the remaining review set. These results indicate that \kgn can narrow the evaluated gap between embedding-based neighborhoods and content-supported semantic relations, while \idp shows that a residual mismatch should not automatically be interpreted as a projection failure.
\end{abstract}

% TODO: Regenerate the final categories with the ACM CCS tool.


\maketitle

\section{Introduction}
\label{sec:introduction}

Embedding visualization is an important analytical tool for exploring similarities and latent structure among data instances in large datasets~\cite{smilkov2016projector,boggust2022comparator,wang2023wizmap,ren2025embeddingatlas}. Analysts commonly interpret nearby instances in a low-dimensional layout as similar or related and use the resulting neighborhoods for data analysis~\cite{liu2024spatialsemantics,molino2019parallax}. Such neighborhood-based interpretation assumes that proximity in the layout reflects relatedness in the underlying content. However, low-dimensional proximity alone does not guaranty semantic relatedness~\cite{liu2026semanticsteering}. Neighborhood-based dimensionality reduction (DR) methods such as UMAP construct a neighborhood graph from the input embedding and optimize a low-dimensional layout to reflect that graph~\cite{mcinnes2018umap}. Consequently, when the embedding-based neighborhood graph does not adequately capture semantic relatedness, the low-dimensional layout may inherit the same limitation. Interpreting proximity in such a layout as evidence of semantic relatedness can therefore mischaracterize the semantic relatedness supported by instance content. This limitation motivates complementing embedding-based neighborhoods with semantic evidence beyond embedding distance alone.

In unsupervised and exploratory settings, external semantic information may be unavailable, whereas information derived solely from the embedding remains limited by what the embedding already represents. External semantic information, such as labels or domain knowledge, can provide explicit guidance~\cite{zhang2026sakura,oliveira2025steerable,liu2026semanticsteering,haisu2022,heiter2023revisedctsne}, but obtaining such information may be costly or infeasible in exploratory analysis. Neighborhood relations can also be refined using the embedding-derived neighborhood structure~\cite{ertoz2003snn,huang2026maple}, yet such refinement remains constrained by information already represented in the input embedding. Instance content provides a complementary source of evidence that depends neither on external semantic information nor solely on the embedding-derived neighborhood structure.

We propose Knowledge-Guided Neighborhood Graph Construction (\kgn), which incorporates content-derived relational knowledge into neighborhood construction without replacing the input embedding. \kgn represents the relational knowledge of each instance as a set of knowledge graph (KG) triples, enabling relational knowledge to be compared across instances. Shared triples provide pairwise semantic evidence that two instances are semantically related even when their embedding distance does not fully reflect that relatedness. \kgn quantifies the degree of triple sharing as KG similarity and uses it to adjust pairwise embedding distances before reconstructing the top-$k$ neighborhood graph. In this way, \kgn complements embedding-based neighborhood construction with content-derived relational knowledge while preserving the input embedding as the underlying representation.

Improving neighborhood construction does not necessarily eliminate every disagreement between an instance and its surrounding neighborhood. Such a disagreement does not by itself imply a projection failure. We use the term \emph{assigned class} to denote the class label provided for an instance in the dataset, without assuming that it is necessarily correct or uniquely supported by the instance content. We define a \emph{residual mismatch} as a condition in which an instance's assigned class differs from its neighborhood-majority class in a low-dimensional layout. This definition applies to every evaluated layout, including the layout obtained from the kNN baseline and those produced after \kgn adjustment. A residual mismatch may indicate that an instance is poorly represented by its neighborhood, but it may also arise because the assigned class is erroneous or because the instance content is semantically ambiguous~\cite{jeon2024classes,northcutt2021labelerrors,plank2022humanvariation}. Treating every residual mismatch as a projection failure would therefore conflate distinct sources of disagreement. Interpreting these cases requires examining whether the instance content supports the assigned class, the neighborhood-majority class, or both.

We perform this instance-level assessment with the \idprofile{} (\idp). For each residual mismatch, \idp independently evaluates whether the instance content supports the assigned class and whether it supports the neighborhood-majority class, and combines these judgments to characterize the disagreement. This assessment is separate from the evidence used for neighborhood construction. \idp does not reuse the KG-based evidence used in \kgn and instead examines the instance content directly. \idp characterizes how the assigned class and the neighborhood-majority class are supported by the instance content, providing a separate basis for interpreting residual mismatches across the evaluated layouts.

We evaluate whether \kgn narrows the gap between the semantic relatedness captured by embedding-based neighborhoods and that supported by instance content, and use \idp to characterize the residual mismatches observed across the evaluated layouts. Across the guidance-strength sweep, \kgn improves both global cluster--class correspondence and local class consistency, with larger gains under encoder conditions with lower kNN baseline scores. The \idp analysis further shows that the instances whose residual mismatches are resolved by neighborhood adjustment differ in composition from those whose mismatches remain, indicating that the remaining mismatches should not automatically be attributed to neighborhood structure alone. Together, these results support a two-stage view of semantic neighborhood analysis: pairwise semantic evidence from content-derived relational knowledge can improve neighborhood structure without replacing the input embedding, while residual mismatches require separate instance-level interpretation rather than being automatically treated as projection failures.

Our main contributions are as follows:
\begin{itemize}
  \item \textbf{Content-derived neighborhood construction:}
  We propose \kgn, which uses shared KG triples derived from instance content as pairwise semantic evidence to adjust embedding distances.

  \item \textbf{Instance-level diagnostic framework:}
  We introduce \idp, which interprets residual mismatches by separately assessing whether the instance content supports the assigned class and whether it supports the neighborhood-majority class.

  \item \textbf{Empirical characterization:}
  We empirically characterize the effects of \kgn across guidance strengths and find that it improves global cluster--class correspondence and local class consistency while partially preserving existing neighborhoods.
\end{itemize}

\section{Related Work}
\label{sec:related-work}

\subsection{Semantic Neighborhoods in Embedding Visualization}
\label{sec:rw-semantic-neighborhoods}

Neighborhood-based DR methods such as UMAP optimize a low-dimensional layout to reflect a neighborhood graph constructed from high-dimensional embeddings~\cite{mcinnes2018umap}. These neighborhoods provide the basis for interpreting relations among instances in embedding visualizations~\cite{smilkov2016projector,boggust2022comparator}. However, depending on the analytical objective, spatial proximity may not correspond to semantic relatedness in the source content~\cite{molino2019parallax,liu2024spatialsemantics,liu2026semanticsteering}. Several methods address this mismatch using external semantic information. Supervised SNE and S-multi-SNE use class labels~\cite{zhang2023supervisedsne,rodosthenous2021smultisne}. Haisu and conditional t-SNE variants incorporate class hierarchies or label structures~\cite{haisu2022,kang2019ctsne,heiter2023revisedctsne}. Steerable projection methods and SAKURA incorporate analyst-provided semantic intent and domain knowledge, respectively~\cite{oliveira2025steerable,liu2026semanticsteering,zhang2026sakura}. These approaches require external semantic information supplied through labels, hierarchies, domain knowledge, or analyst interaction. Their applicability is therefore limited when such information is unavailable.

Another line of research refines neighborhood relations using only the input embedding and structures derived from it. SNN redefines pairwise similarity based on shared nearest neighbors~\cite{ertoz2003snn}. LocalMAP dynamically removes likely false-positive neighbor edges and adds further-pair edges during layout optimization~\cite{wang2025localmap}. MAPLE uses initial neighborhood relations as self-supervised signals to learn a new embedding space and reconstruct the neighborhood graph~\cite{huang2026maple}. These methods do not require external semantic information, but their refinement signals remain tied to the input embedding or its initial neighborhood graph. They do not explicitly construct pairwise semantic evidence from content-derived relational knowledge. \kgn instead derives relational knowledge from the content of each instance and uses shared KG triples as pairwise semantic evidence. This evidence complements embedding distance during neighborhood construction.

\subsection{Interpreting Residual Mismatch}
\label{sec:rw-mismatch}

Existing approaches to projection evaluation measure structural discrepancies between input and low-dimensional spaces. Trustworthiness, continuity, and local visualization methods assess false and missing neighbors~\cite{venna2010information,colange2019interpreting}. Local label trustworthiness and qHD examine label mismatches and class boundaries~\cite{venna2006lmds,huttenlocher1993hausdorff,taha2015metrics,huang2026maple}. Label-Trustworthiness and Label-Continuity measure changes in cluster--class correspondence~\cite{jeon2024classes}. These methods characterize neighborhood preservation, class mixing, and class separation. However, these evaluations do not by themselves indicate how to interpret a mismatch between the assigned class of a particular instance and its neighborhood-majority class. Detecting or locating such a mismatch is distinct from interpreting it in light of the source content.

This interpretive problem also arises in research on label validity. Label-error studies assume a latent ground-truth label distinct from the observed label~\cite{northcutt2021labelerrors,northcutt2021confident}. In contrast, research on human label variation shows that instance ambiguity and multiple interpretations can make more than one label valid~\cite{plank2022humanvariation,pavlick2019disagreements}. A residual mismatch alone is therefore insufficient to determine which class should be considered correct. Its nature is also difficult to assess in single-label datasets that lack repeated human judgments or annotator-level labels. For each detected residual mismatch, \idp separately evaluates whether the source content supports the assigned class and whether it supports the neighborhood-majority class. It combines these judgments to interpret each mismatch at the instance level.

\section{Method}
\label{sec:method}

\subsection{Overview}
\label{sec:method-overview}

We propose \kgn to guide neighborhood graph construction using semantic evidence derived from instance content and introduce \idp as a supporting framework for diagnosing residual mismatches in each evaluated layout. Let $\mathcal{X}=\{x_i\}_{i=1}^{N}$ denote a dataset of $N$ instances, where $x_i$ is the $i$th instance. Each $x_i$ has a $d$-dimensional input embedding $\mathbf{z}_i\in\mathbb{R}^{d}$. \kgn adjusts pairwise embedding distances using the similarity between instance-specific KG triple sets and reconstructs the top-$k$ neighbors from the adjusted distances. Importantly, \kgn modifies only neighborhood construction, leaving the input embeddings and encoder unchanged. Subsequent graph weighting and low-dimensional layout optimization follow the original DR procedure.

\subsection{Content-Derived Relational Knowledge}
\label{sec:knowledge-semantic-evidence}

To compare relational knowledge across instances, we represent the entities and relations in each $x_i$ as a set $T_i$ of subject--predicate--object triples. We construct a separate triple set for each instance instead of integrating all instances into one KG. These sets provide a common intermediate representation for comparing relational knowledge derived from instance content.

Exact triple matching requires consistent representations that remain faithful to the source content. Drawing on prior work on staged decomposition and agent collaboration in LLM-based KG construction~\cite{carta2023iterativekg,lairgi2025itext2kg,chen2024sackg,ye2023cooperkg}, as well as schema normalization and source verification~\cite{zhang2024edc,han2024pive,kim2026graphrefine}, we organize KG construction into Entity Extraction, Relation Linking, and Critique. The workflow normalizes explicitly stated entities, links them with schema-defined predicates, and checks the candidate triples against the source. It removes erroneous and duplicate triples and adds only relations explicitly supported by the content. The domain schema constrains the representation vocabulary but does not specify proximity between instances, class labels, or class hierarchies. Triples shared by $T_i$ and $T_j$ provide pairwise semantic evidence of relatedness between the corresponding instances. Section~\ref{sec:implementation-details} describes the modality-specific inputs and extraction model.

\subsection{Knowledge-Guided Neighborhood Graph Construction}
\label{sec:kg-guided}

To incorporate semantic evidence from instance-specific relational knowledge into neighborhood construction, \kgn computes KG similarity for each pair of instances. It uses this similarity to adjust pairwise embedding distances and then reconstructs the top-$k$ neighbors from the adjusted distances.

\textbf{Knowledge Graph Similarity Calculation.}
We compute KG similarity $\Skg(i,j)\in[0,1]$ between triple sets $T_i$ and $T_j$. We compare four similarity measures to examine how the choice of measure affects the results.

Jaccard similarity normalizes the number of shared triples by the size of the union of the two sets~\cite{kosub2019jaccard}, whereas the overlap coefficient normalizes it by the size of the smaller set to capture cases in which one KG is largely contained in the other.

\begin{gather}
S_{\mathrm{Jaccard}}(i,j)
= \frac{|T_i \cap T_j|}{|T_i \cup T_j|},
\qquad
S_{\mathrm{overlap}}(i,j)
= \frac{|T_i \cap T_j|}{\min(|T_i|,|T_j|)}.
\label{eq:jaccard-overlap}
\end{gather}

IDF-weighted cosine similarity and normalized Adamic--Adar similarity reduce the influence of common triples that occur across many instances. Let $\mathrm{df}(t)$ denote the number of instances in which triple $t$ appears. IDF-weighted cosine similarity computes the cosine similarity between vectors that assign greater weight to rare triples~\cite{sparckjones1972idf}.

\begin{equation}
S_{\mathrm{idf}}(i,j)
= \frac{\mathbf{v}_i^\top\mathbf{v}_j}
{\|\mathbf{v}_i\|_2\|\mathbf{v}_j\|_2},
\qquad
\mathbf{v}_i[t]
=\log\!\left(\frac{N}{\mathrm{df}(t)}\right)
\mathbb{1}[t\in T_i].
\label{eq:idf-cosine}
\end{equation}

Adamic--Adar similarity emphasizes rare shared triples by summing the inverse logarithm of their occurrence frequencies~\cite{adamic2003friends} and is normalized by the maximum value over all instance pairs.

\begin{equation}
S_{\mathrm{AA}}(i,j)
= \frac{A(i,j)}{\max_{p\neq q}A(p,q)},
\quad
A(i,j)
=\sum_{t\in T_i\cap T_j}
\bigl(\log \mathrm{df}(t)\bigr)^{-1}.
\label{eq:adamic-adar}
\end{equation}

We set undefined similarities and diagonal entries corresponding to self-comparisons to zero. Because all four measures use exact triple matching, they do not directly account for partial entity or relation matches, lexical similarity, or multi-hop graph structure.

\textbf{Knowledge-Guided Distance Adjustment.}
To incorporate KG similarity into neighbor selection, \kgn adjusts the pairwise embedding distances. Let $\mathbf{z}_i$ and $\mathbf{z}_j$ be the embeddings of instances $x_i$ and $x_j$. Their cosine distance is

\begin{equation}
\Demb(i,j)
=1-\frac{\mathbf{z}_i^\top\mathbf{z}_j}
{\|\mathbf{z}_i\|_2\|\mathbf{z}_j\|_2}.
\label{eq:embedding-distance}
\end{equation}

We use $\Skg(i,j)\in[0,1]$ to reduce the embedding distance between instances with high KG similarity. The KG-adjusted embedding distance is

\begin{equation}
\Dkg(i,j)
= \Demb(i,j)\bigl(1-\Skg(i,j)\bigr).
\label{eq:kg-adjusted-distance}
\end{equation}

Here, $\Dkg$ is not a distance computed independently in a separate KG space but the original embedding distance adjusted by KG similarity. We combine it with $\Demb$ to define the knowledge-guided distance. We refer to $\alpha\in[0,1]$ as the \emph{guidance strength}, which controls the influence of KG similarity on the embedding distance.

\begin{equation}
\Dkgn(i,j)
= (1-\alpha)\Demb(i,j)+\alpha\Dkg(i,j),
\qquad \alpha \in [0,1].
\label{eq:kg-guided-distance}
\end{equation}

Substituting Equation~\ref{eq:kg-adjusted-distance} into Equation~\ref{eq:kg-guided-distance} gives the equivalent form $\Dkgn(i,j)=\Demb(i,j)\bigl(1-\alpha\Skg(i,j)\bigr)$.
When $\Skg(i,j)=0$, the original distance is preserved. When $\Skg(i,j)=1$, the distance decreases to $(1-\alpha)\Demb(i,j)$. Setting $\alpha=0$ yields the same neighborhood as the original kNN construction.

\begin{figure}[t]
  \centering
  \includegraphics[width=\columnwidth]{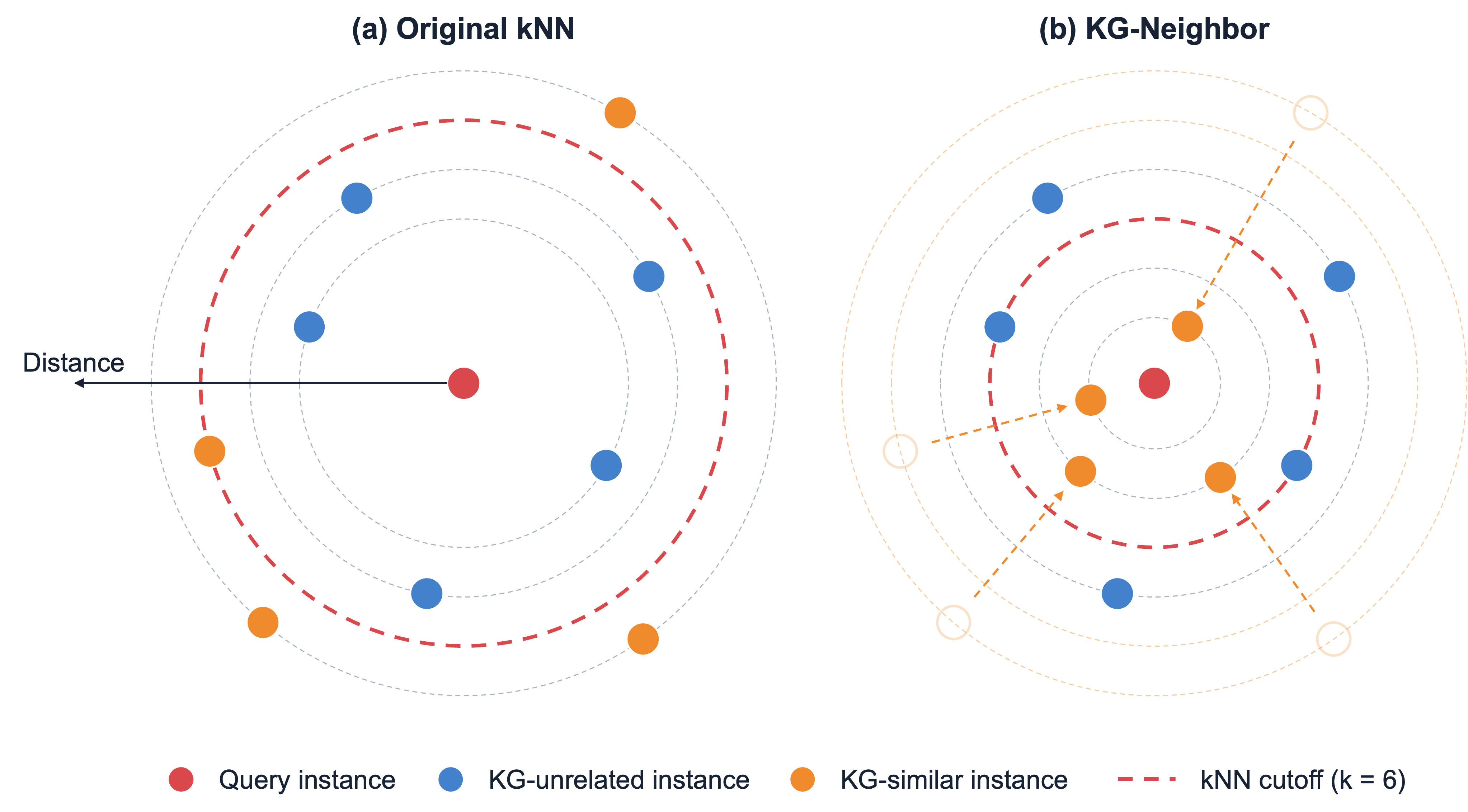}
  \caption{Comparison of (a) kNN and (b) \kgn. Knowledge-guided distance adjustment reduces the effective distances between the query and KG-similar instances, which changes the top-$k$ neighbor set. The input embeddings remain unchanged.}
  \Description{A two-panel illustration compares kNN with KG-Neighbor for one query instance. KG-similar instances move inside the top-six cutoff after their effective distances are reduced, while KG-unrelated instances retain their original distances.}
  \label{fig:kg-guided-knn}
\end{figure}

For each instance $x_i$, we select the $k$ instances $x_j$ with the smallest $\Dkgn(i,j)$ values among all $j\neq i$ as its new neighbors. Although this multiplicative distance modulation is mathematically similar to hierarchy-guided distance adjustment~\cite{haisu2022}, \kgn computes each pairwise adjustment from triples shared by the two instance-specific KGs rather than from a predefined class hierarchy. Figure~\ref{fig:kg-guided-knn} illustrates how KG similarity changes the top-$k$ neighbors without modifying the input embeddings.

\subsection{Instance-Level Diagnosis of Residual Mismatch}
\label{sec:idp}

Residual neighborhood--class mismatches may occur in both the kNN baseline layout at $\alpha=0$ and layouts produced after neighborhood adjustment using semantic evidence. \idp detects these instances and interprets each mismatch by separately assessing whether the source content supports the assigned class and whether it supports the neighborhood-majority class.

\textbf{Residual Mismatch Detection.}
For layout $r$, let $\mathbf{y}_i^{(r)}\in\mathbb{R}^{q}$ denote the $q$-dimensional coordinates of instance $i$, and let $N_k^{(r)}(i)$ denote its Euclidean $k$-nearest-neighbor set excluding the instance itself. Let $A_i$ denote the assigned class, and let $\mathcal{C}$ denote the set of all classes. The number of neighbors belonging to each class and the neighborhood-majority class are defined as

\begin{equation}
n_i^{(r)}(c)
=\sum_{j\in N_k^{(r)}(i)}\mathbb{1}[A_j=c],
\qquad
B_i^{(r)}
=\underset{c\in\mathcal{C}}{\arg\max}\;n_i^{(r)}(c).
\label{eq:neighborhood-modal-label}
\end{equation}

We detect a residual mismatch for instance $i$ when $A_i\neq B_i^{(r)}$ and denote the set of detected instances by $\mathcal{M}_r$. If multiple classes tie for the maximum neighbor count, we select the tied class with the smallest mean distance, measured in the low-dimensional layout, from instance $i$ to the neighbors assigned to that class. If the mean distances are also equal, we follow a predefined class ordering. Detection is performed independently for each layout with $q=2$. A detected instance is only a candidate for further review and does not by itself indicate an error in either the class or the layout.

\textbf{Instance-Level Judgment.}
For each detected residual mismatch, we assess whether the source content supports each of the two candidate classes. We define $J(x_i,c;\mathcal{C})$ as a binary judgment function indicating whether the content of instance $x_i$ supports the target class $c$ given the set of all classes $\mathcal{C}$. The judgments for the assigned class and the neighborhood-majority class are

\begin{equation}
a_i=J(x_i,A_i;\mathcal{C}),
\quad
b_i^{(r)}=J(x_i,B_i^{(r)};\mathcal{C}),
\quad
a_i,b_i^{(r)}\in\{0,1\}.
\label{eq:idp-judgments}
\end{equation}

We make the two judgments independently without using information from the low-dimensional layout.

We distinguish four diagnostic states based on the two judgments. We interpret $a_i=1$ and $b_i^{(r)}=0$ as a \emph{Placement Issue}. We interpret $a_i=0$ and $b_i^{(r)}=1$ as a \emph{Label Issue}. If the source content supports both classes, then $a_i=b_i^{(r)}=1$ and the instance is a \emph{Boundary Case}. If the source content supports neither class, then $a_i=b_i^{(r)}=0$ and the instance is an \emph{Unresolved Case}. These states do not establish the cause of a mismatch. They characterize how the assigned class and the neighborhood-majority class relate to the source content.

\idp reports the detection rate and the absolute number of instances in each diagnostic state. The detection rate is the number of mismatches in a layout divided by the total number of instances. We implement the judgment function with an LLM/VLM-based judge. Section~\ref{sec:implementation-details} provides the configuration.

\section{Experiments}
\label{sec:experiments}

We evaluate how semantic evidence derived from instance content complements embedding-based neighborhood construction and what the resulting residual mismatches indicate. The evaluation addresses three research questions.

\begin{itemize}
  \item \textbf{RQ1.} How does semantic proximity in the low-dimensional layout vary across KG similarity measures and guidance strengths when semantic evidence derived from instance content is incorporated into neighborhood construction? (Figures~\ref{fig:alignment-alpha-sweep} and~\ref{fig:qualitative-umap})
  \item \textbf{RQ2.} How do changes in label-based semantic proximity differ between encoder conditions with lower and higher kNN baseline scores, and how are they related to neighborhood reconfiguration across guidance strengths? (Figures~\ref{fig:alignment-alpha-sweep} and~\ref{fig:neighbor-retention})
  \item \textbf{RQ3.} How do the scale and diagnostic-state composition of residual mismatches in the low-dimensional layout vary across guidance strengths? (Figure~\ref{fig:idp-alpha-sweep})
\end{itemize}

\subsection{Experimental Setup}
\label{sec:experimental-setup}

\textbf{Datasets.} We use the text datasets StackOverflow~\cite{xu2015stackoverflow} and BANKING77~\cite{casanueva2020banking77}, which provide source text for relational knowledge extraction. For images, we use CUB-200-2011~\cite{wah2011cub} and Oxford Flowers 102~\cite{nilsback2008flowers}, for which detailed visual descriptions are available~\cite{reed2016descriptions}. To control for differences in class count across datasets, we sample 20 classes from each dataset using a random seed of 0 and retain all instances in the selected classes. Because StackOverflow contains exactly 20 classes, we use the full dataset. For BANKING77, we combine the public training, validation, and test splits before sampling classes (Table~\ref{tab:datasets}).

\begin{table}[t]
  \caption{Evaluation subsets and mean KG triples per instance. $C_{\mathrm{all}}$ and $C$ denote the original and retained numbers of classes.}
  \label{tab:datasets}
  \centering
  \small
  \setlength{\tabcolsep}{5pt}
  \begin{tabular}{lcrrrr}
    \toprule
    Dataset & Modality & $C_{\mathrm{all}}$ & $C$ & $N$ & Triples/inst. \\
    \midrule
    StackOverflow & text  & 20  & 20 & 20,000 & 6.1  \\
    BANKING77     & text  & 77  & 20 & 3,256  & 6.4  \\
    CUB-200-2011  & image & 200 & 20 & 1,182  & 22.8 \\
    Flowers 102   & image & 102 & 20 & 1,384  & 15.9 \\
    \bottomrule
  \end{tabular}
\end{table}

\textbf{Metrics.} Class labels do not fully define semantic relatedness; we use them only as an evaluation reference for measuring global cluster--class correspondence and local class consistency. For RQ1, we measure global cluster--class correspondence using normalized mutual information (NMI) between $k$-means assignments ($K=C$) and dataset labels~\cite{vinh2009information}. We do not interpret NMI as a measure of the layout's overall geometric quality~\cite{jeon2024classes}. We measure local class consistency using rank-weighted local label trustworthiness ($\mathrm{LT}_{2D}$) on the 15-NN graph constructed in the two-dimensional layout~\cite{venna2006lmds,huang2026maple}. For RQ2, we compare the gaps in NMI and $\mathrm{LT}_{2D}$ between the weak and stronger encoder conditions across guidance strengths. We also report Neighbor Retention, defined as the mean fraction of original top-$k$ neighbors retained after distance adjustment. For RQ3, we report the residual mismatch detection rate and the number of instances in each of the four diagnostic states defined in Section~\ref{sec:idp}. Class labels are used only for evaluation and \idp diagnosis. They are not provided during KG extraction or \kgn neighborhood construction.

\textbf{Baselines.} We use \textbf{kNN} ($\alpha=0$) as the reference condition; its neighborhood graph is constructed directly from embedding distances. We compare \kgn with \textbf{SNN}~\cite{ertoz2003snn}, which redefines pairwise similarity based on the number of shared nearest neighbors in the input embedding, and \textbf{MAPLE}~\cite{huang2026maple}, which learns a new representation through self-supervised training. We exclude Haisu~\cite{haisu2022} because it relies on class hierarchies. Applying it in our nonhierarchical setting would expose the flat labels used for evaluation during neighborhood construction.

\subsection{Implementation Details}
\label{sec:implementation-details}

\textbf{Inputs and Representations.} We use Claude Opus 4.8 to construct KGs without providing class information. For text datasets, the model receives the original text of each instance; for image datasets, it receives ten detailed visual descriptions of each image~\cite{reed2016descriptions}. Because exact triple matching requires consistent representations, we apply the same domain-specific entity types, predicate sets, and normalization rules throughout Entity Extraction, Relation Linking, and Critique. The KG schemas and prompts used in the three KG-construction stages are provided in the supplementary material. For the input embeddings, we use \texttt{bert-base-}\allowbreak\texttt{uncased}~\cite{devlin2019bert} and Qwen3-Embedding-8B~\cite{zhang2025qwen3embedding} for text, and ResNet-50~\cite{he2016resnet} and CLIP ViT-L/14~\cite{radford2021clip} for images. All encoders use pretrained weights without additional training, and their embeddings are $\ell_2$-normalized. To compare the effect of \kgn across different baseline conditions, we distinguish the two encoder conditions for each dataset based on their kNN baseline NMI and $\mathrm{LT}_{2D}$. We refer to the lower- and higher-performing conditions as \emph{weak} and \emph{stronger}, respectively. These terms indicate relative performance within each dataset rather than general encoder quality.

% We refer to the encoder conditions with lower and higher NMI and $\mathrm{LT}_{2D}$ under the kNN baseline as \emph{weak} and \emph{stronger}, respectively. These terms reflect performance in this evaluation rather than a general ranking of encoder capacity.

\textbf{Neighborhood and Projection Settings.} We use $k=15$ and cosine distance for kNN, SNN, and \kgn. We evaluate all four KG similarity measures at each guidance strength $\alpha\in\{0,.1,.2,\ldots,1\}$. Setting $\alpha=0$ reproduces kNN. The quantitative analysis uses the full sweep to characterize sensitivity to guidance strength rather than to select a condition-specific setting. For the visual comparison, we use kNN ($\alpha=0$) and \kgn at $\alpha\in\{.3,.7,1\}$ as anchor settings spanning the evaluated range. For kNN, SNN, and \kgn, we use UMAP to generate two-dimensional layouts~\cite{mcinnes2018umap} and apply the same graph-weighting and layout-optimization procedures after neighborhood construction. We configure UMAP with \texttt{min\_dist}=0.1, 200 epochs, spectral initialization, and a random seed of 0. We use the default settings provided by the public implementations of SNN and MAPLE.

\textbf{Judge Configuration.} We use an LLM to assess whether the instance content supports each candidate class in \idp~\cite{zheng2023judge}. We implement the judgment function with GPT-5.4 and set the temperature to 0 and reasoning effort to \texttt{none} to ensure deterministic and reproducible judgments. Because an LLM may select a provided candidate even when none is appropriate~\cite{xu2024overclaimed}, we do not ask the model to choose between the assigned class and the neighborhood-majority class. Instead, we evaluate each class in a separate binary request. This design allows the instance content to support both classes or neither class. Each request provides only the instance content, the target class, and the complete class list; no layout information is included. We reuse each judgment for the same instance--class pair across all layouts. The system and user prompts used for the \idp judgments are provided in the supplementary material.

\section{Results}
\label{sec:results}

We first evaluate how evidence derived from instance content affects semantic proximity in the low-dimensional layout. We then examine how this effect varies across input-embedding conditions and with the extent of neighborhood reconfiguration. Finally, we use \idp to analyze how the scale and diagnostic-state composition of residual mismatches vary across guidance strengths.

% We evaluate \kgn with respect to three research questions. RQ1 asks whether \kgn improves global and local class alignment in the low-dimensional visualization, RQ2 examines the extent to which these improvements occur while retaining existing neighbors, and RQ3 analyzes how the scale and composition of residual mismatch change after adjustment. Unless otherwise noted, our representative setting uses the overlap coefficient at $\alpha=0.7$, the smallest common guidance strength at which both metrics exceed the kNN baseline in every dataset--encoder combination. We distinguish results under this representative setting from the combination-specific peak performance obtained over the parameter sweep.

\subsection{RQ1: Changes in Semantic Proximity}
\label{sec:semantic-proximity-results}

\begin{figure*}[t]
  \centering
  \includegraphics[width=\textwidth]{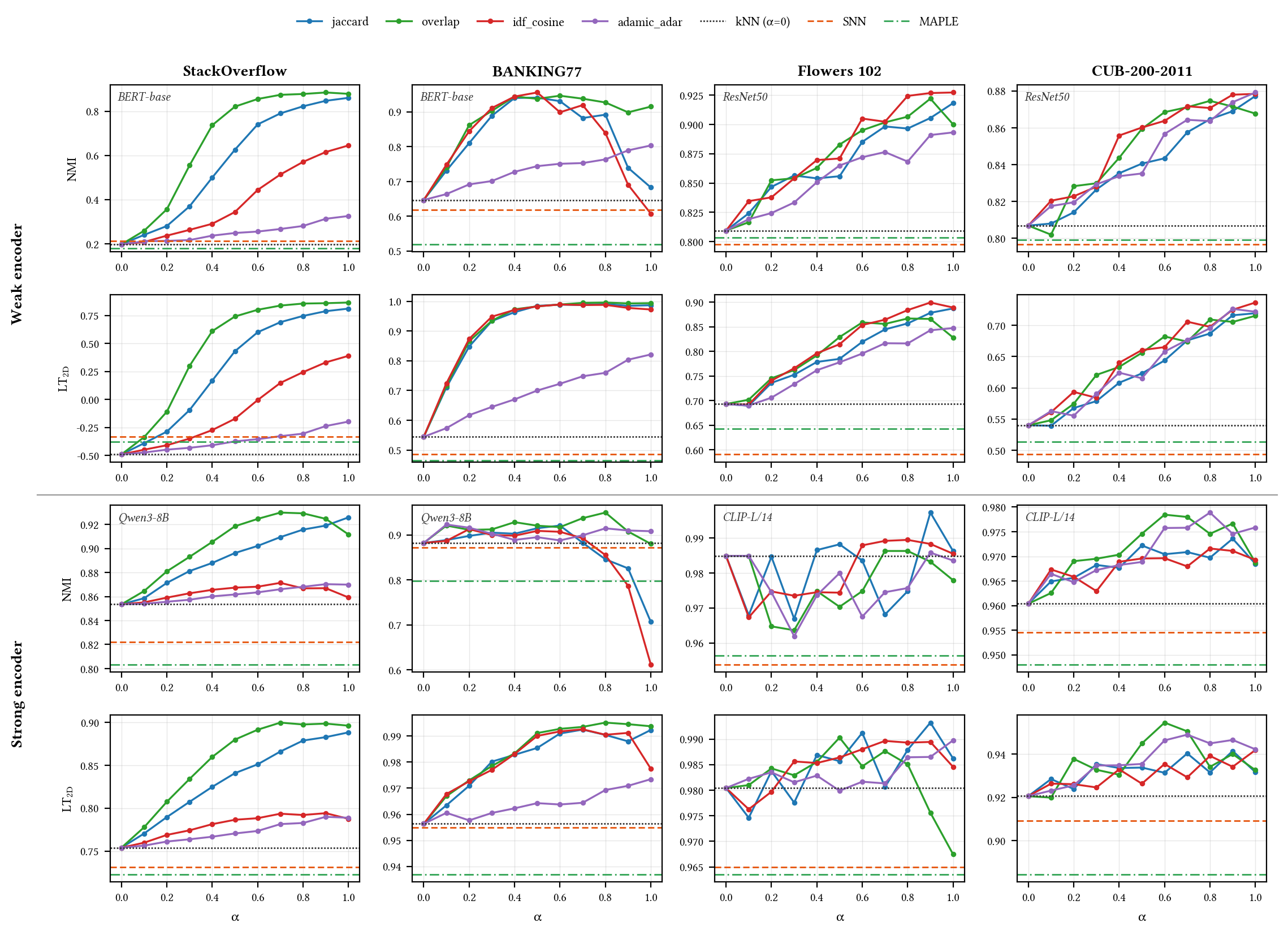}
  \caption{Global cluster--class correspondence and local class consistency across guidance strengths and KG similarity measures. The first two rows show the weak encoder conditions, and the last two show the stronger encoder conditions; within each pair, the first row reports NMI and the second reports rank-weighted local label trustworthiness ($\mathrm{LT}_{2D}$). The dotted, dashed, and dash-dotted lines denote the kNN ($\alpha=0$), SNN, and MAPLE baselines.}
  \Description{Sixteen line charts compare NMI and local label trustworthiness as guidance strength increases from zero to 1.0 for StackOverflow, BANKING77, Flowers 102, and CUB-200-2011. Four curves compare Jaccard, overlap coefficient, IDF-weighted cosine, and normalized Adamic--Adar similarity, while three horizontal lines mark the kNN, SNN, and MAPLE baselines.}
  \label{fig:alignment-alpha-sweep}
\end{figure*}

\begin{figure*}[t]
  \centering
  \includegraphics[width=\textwidth]{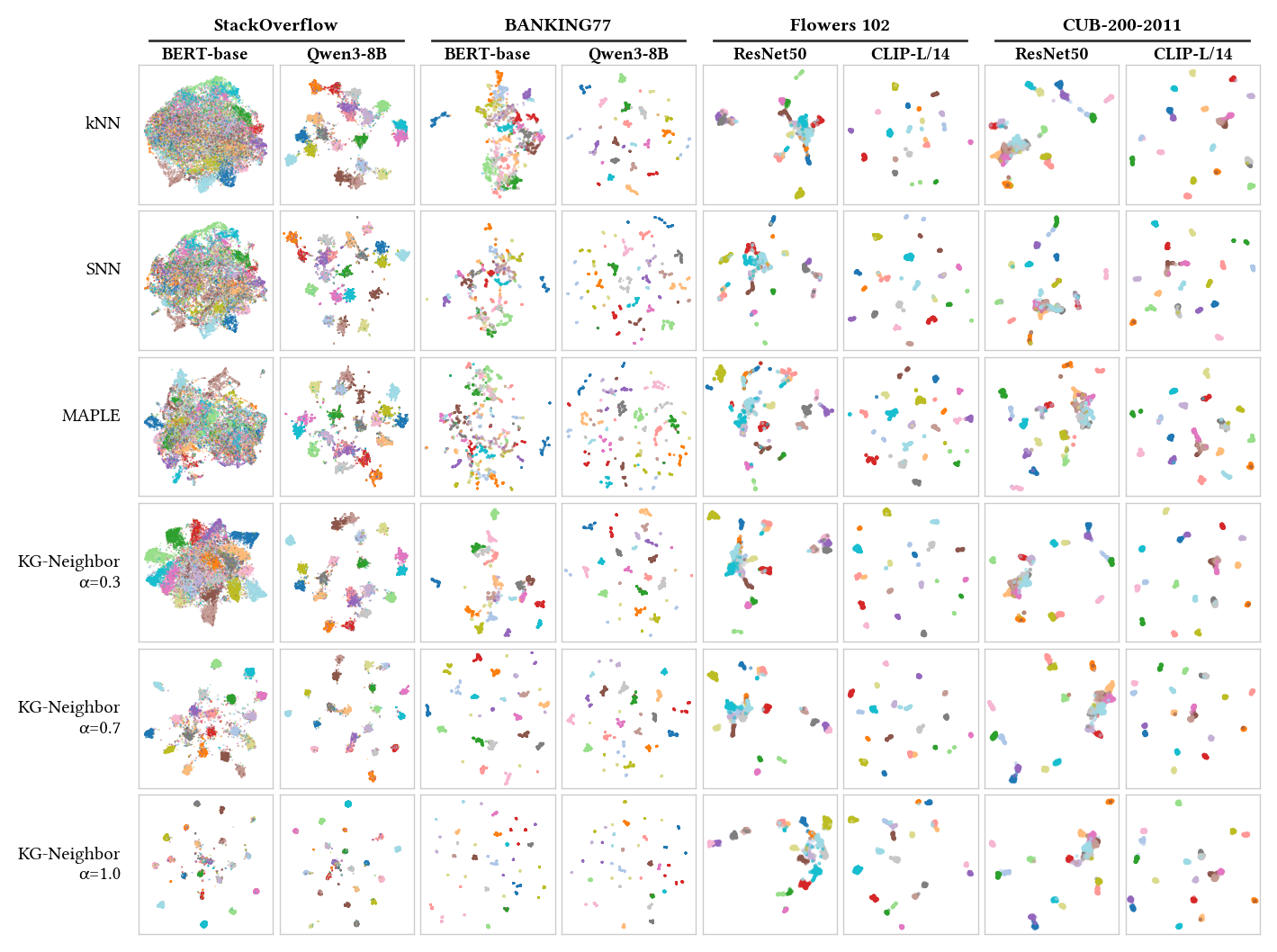}
  \caption{Qualitative comparison of UMAP layouts across datasets and encoders. Rows compare kNN, SNN, MAPLE, and \kgn with $\alpha\in\{.3,.7,1\}$ using the overlap coefficient. Colors denote dataset labels used only for evaluation.}
  \Description{A grid of 48 UMAP projections compares four datasets, each with a weak and a stronger encoder condition. Six rows show kNN, SNN, MAPLE, and three knowledge-guided conditions, with points colored by class.}
  \label{fig:qualitative-umap}
\end{figure*}

Figure~\ref{fig:alignment-alpha-sweep} reports the quantitative results across all KG similarity measures and guidance strengths. The trajectories vary across KG similarity measures and dataset--encoder conditions, with broader improvements under the weak encoder conditions and smaller or non-monotonic changes under several stronger encoder conditions. For every KG similarity measure and dataset--encoder combination, the maximum value of each metric over the guidance-strength sweep exceeds the corresponding kNN, SNN, and MAPLE baselines. The improvement in these label-based measures is therefore not confined to a particular KG similarity measure or encoder condition, although its magnitude varies across input embeddings. For the overlap coefficient, the mean gains of the condition-specific maxima over kNN are 0.2920 in NMI and 0.5393 in $\mathrm{LT}_{2D}$ under the weak encoder conditions. The corresponding gains under the stronger encoder conditions are 0.0409 and 0.0571. The weak encoder conditions thus show large gains in both global cluster--class correspondence and local class consistency. The gains are smaller under the stronger encoder conditions, but local class consistency still improves. Based on this comparison, we use the overlap coefficient in the subsequent analyses because it yields the largest mean peak gain in both metrics.

Figure~\ref{fig:qualitative-umap} shows that variation in metric gains across encoder conditions corresponds to distinct patterns of layout change. Across the StackOverflow--BERT-base layouts, instances from classes that are widely mixed under the kNN baseline appear progressively more localized as guidance strength increases from \(\alpha=0\) to \(1\), with increasingly distinct class-specific regions. By contrast, the broad separation patterns already visible in the StackOverflow--Qwen3-8B and Flowers 102--CLIP-L/14 baselines remain largely unchanged, although $\mathrm{LT}_{2D}$ increases. The weak encoder conditions therefore show changes in both the visible layout organization and local class consistency. The stronger conditions retain their broad organization while improving local class consistency.

The full guidance-strength sweep shows that evidence derived from instance content can improve global cluster--class correspondence and local class consistency in the evaluated dataset--encoder conditions. The magnitude and visible form of the change depend on the KG similarity measure, guidance strength, and input-embedding condition. These findings identify neighborhood construction as an intervention point within the neighborhood-based DR pipeline, but they do not establish an improvement in overall projection quality because dataset labels do not exhaustively define semantic relatedness. The larger gains under weak encoder conditions suggest that content-derived evidence may compensate for differences between input embeddings. RQ2 therefore examines the reduction in metric gaps between encoder conditions together with the extent of neighborhood reconfiguration.

\subsection{RQ2: Gaps Between Encoder Conditions and Neighborhood Reconfiguration}
\label{sec:neighborhood-reconfiguration-results}

\begin{figure*}[t]
  \centering
  \includegraphics[width=\textwidth]{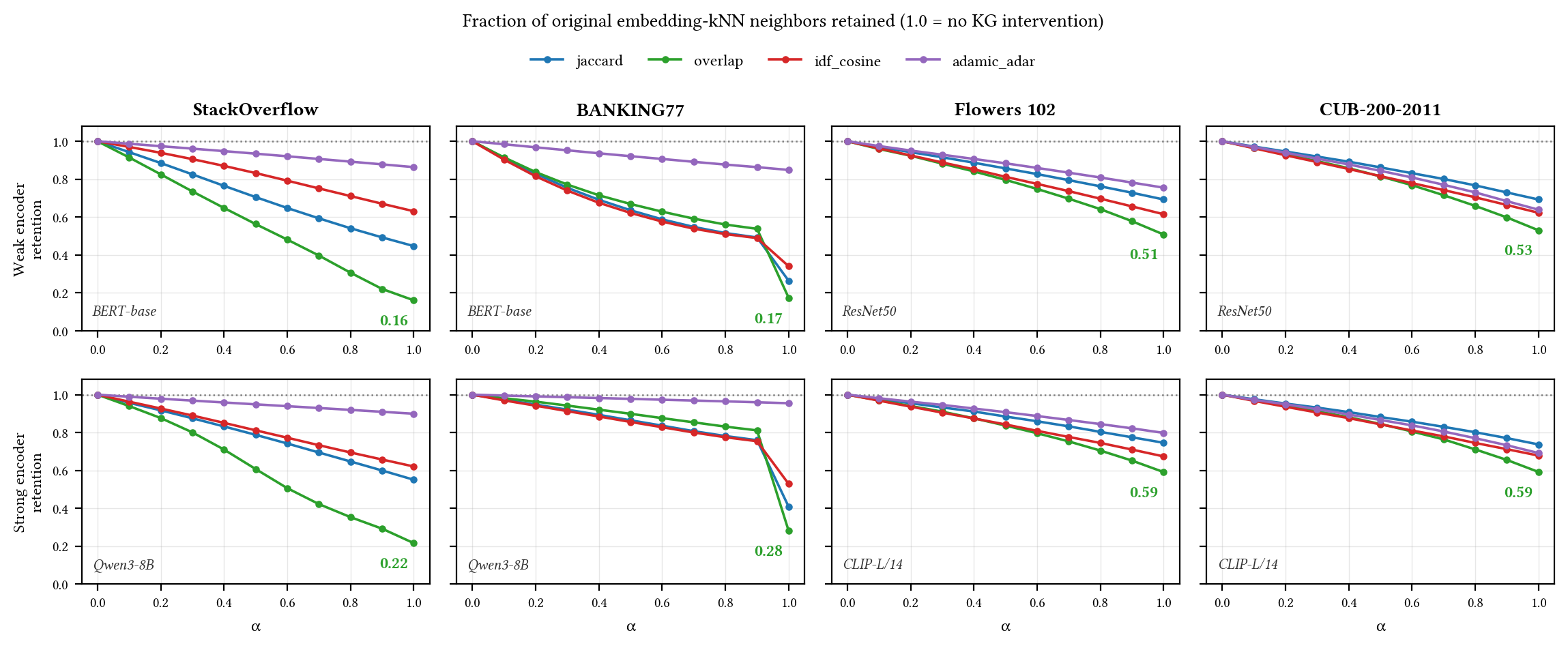}
  \caption{Retention of original kNN neighbors across guidance strengths. A value of 1.0 indicates that the original top-$k$ set is unchanged; lower values indicate a larger change in neighborhood composition. Each panel annotates the lowest retention at $\alpha=1$.}
  \Description{Eight line charts show the fraction of original kNN neighbors retained as guidance strength increases across four datasets. The top row shows the weak encoder conditions, and the bottom row shows the stronger encoder conditions. Four curves compare the KG similarity measures.}
  \label{fig:neighbor-retention}
\end{figure*}

We examine how the gap between the weak and stronger encoder conditions varies across guidance strengths under the overlap coefficient. At $\alpha=0$, the mean gaps across the four datasets are 0.305 in NMI and 0.580 in $\mathrm{LT}_{2D}$. As guidance strength increases, the weak encoder conditions generally show larger increases in both metrics than the stronger conditions, thereby narrowing the gaps between the encoder conditions. This convergence is most pronounced for StackOverflow and BANKING77, where the weak-encoder \kgn scores reach or exceed the stronger encoder's kNN baselines at multiple guidance strengths. The gaps also narrow for Flowers 102 and CUB-200-2011, but residual differences between the encoder conditions remain. At higher guidance strengths, several trajectories plateau or become non-monotonic. The reduction in encoder gaps should therefore be interpreted as a sensitivity pattern that varies across datasets and input embeddings rather than as an effect tied to a particular guidance strength.

Figure~\ref{fig:neighbor-retention} relates these metric changes to neighborhood reconfiguration across the guidance-strength sweep. Neighbor Retention progressively decreases with guidance strength across the dataset--encoder conditions, indicating that larger fractions of the original kNN neighbors are replaced after distance adjustment. At the same guidance strength, the weak encoder conditions generally exhibit higher replacement rates than their corresponding stronger conditions. Within each dataset, the weak encoder condition generally exhibits both more extensive neighborhood reconfiguration and larger metric gains than the corresponding stronger condition. However, continued increases in the replacement rate at higher guidance strengths are not accompanied by proportional improvements in NMI or $\mathrm{LT}_{2D}$. In several conditions, the metric trajectories plateau or reverse while neighborhood reconfiguration continues to increase. The extent of neighborhood reconfiguration alone therefore does not account for the observed metric changes.

Taken together, these results indicate that \kgn yields larger gains in NMI and $\mathrm{LT}_{2D}$ and more extensive neighborhood reconfiguration under the weak encoder conditions. This pattern should not be interpreted as showing that content-derived evidence can replace the stronger encoder conditions evaluated here. For the text datasets, applying \kgn to the weak encoder reaches or exceeds the stronger encoder's kNN baseline at multiple guidance strengths, whereas gaps between the encoder conditions remain for the image datasets. Content-derived evidence should therefore be interpreted as complementary information that can selectively supplement neighborhood relations not fully reflected in the input embedding, rather than as a substitute for stronger input representations. Because improvements in aggregate semantic-proximity measures and reductions in encoder gaps do not imply that every instance-level mismatch has been resolved, RQ3 examines how the scale and composition of residual mismatches vary with guidance strength.

\subsection{RQ3: Residual Mismatch Analysis}
\label{sec:residual-mismatch-results}

\begin{figure*}[t]
  \centering
  \includegraphics[width=\textwidth]{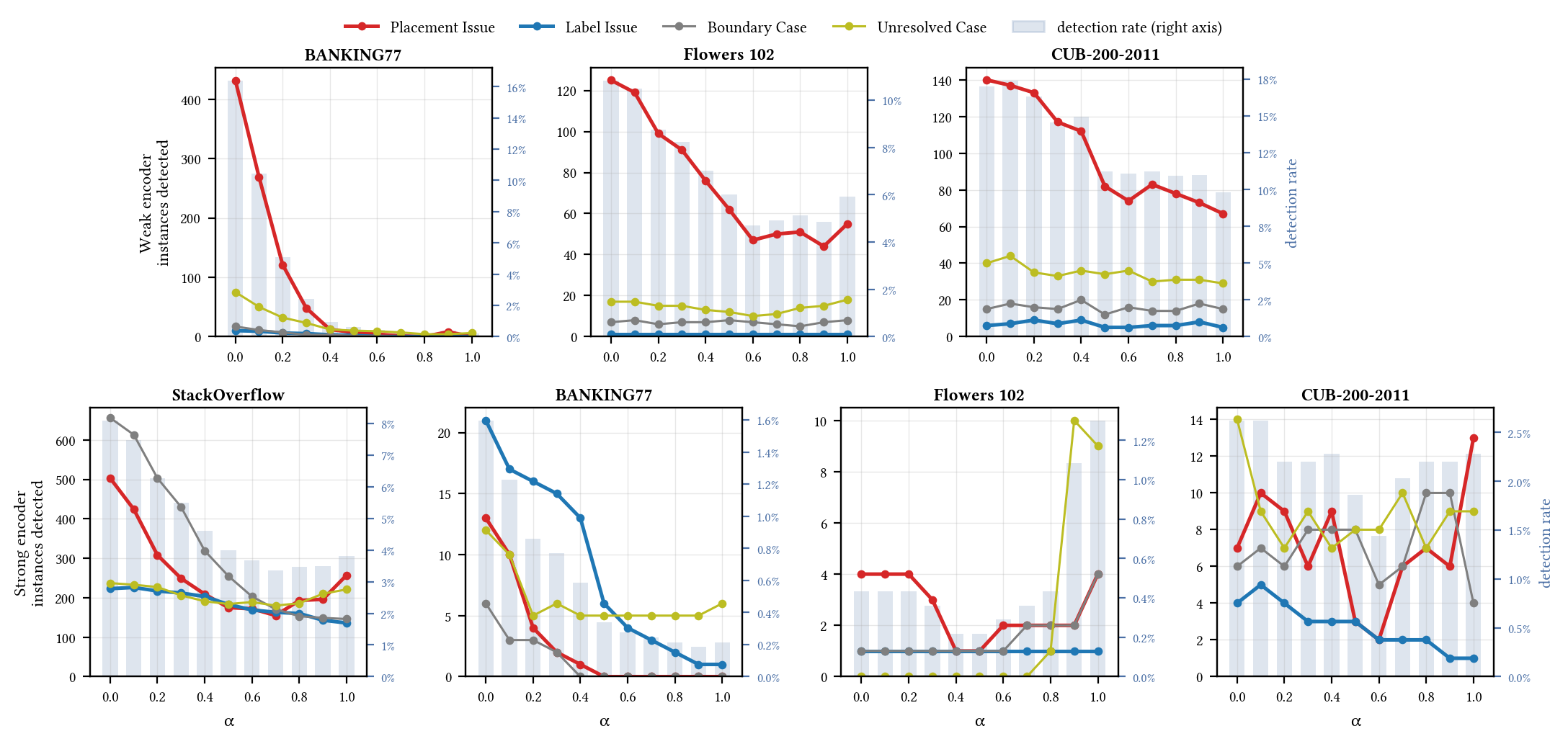}
  \caption{Residual mismatch detection rates and state-specific instance counts across guidance strengths under the overlap coefficient. Curves show the number of instances in each diagnostic state, and shaded bars show the detection rate.}
  \Description{Seven panels show residual mismatch results across guidance strengths. Four curves in each panel report state-specific instance counts, and shaded bars report the detection rate.}
  \label{fig:idp-alpha-sweep}
\end{figure*}

Figure~\ref{fig:idp-alpha-sweep} reports how residual mismatch detection rates vary across guidance strengths under the overlap coefficient. We exclude Stack-Overflow--BERT-base because more than 60\% of its instances are already detected at $\alpha=0$, yielding a baseline candidate set that is insufficiently selective. Across the remaining seven dataset--encoder conditions, detection rates remain below the kNN baselines over much of the guidance-strength sweep. The reductions are more pronounced in conditions with relatively large baseline candidate sets, whereas the absolute changes are smaller in stronger encoder conditions whose baseline detection rates are already low. These results indicate that neighborhood adjustment using content-derived evidence can reduce the number of residual mismatch candidates requiring subsequent diagnosis. The detection rate merely describes the size of the candidate set and does not distinguish which diagnostic states decrease or remain.

The state-specific trajectories show that the reduction in residual mismatch candidates is not uniform across the four diagnostic states. Placement Issue candidates exhibit the most pronounced decreases in conditions with sufficiently large baseline candidate sets. Boundary Case candidates decrease in some conditions but remain stable or fluctuate in others. By contrast, Label Issue and Unresolved Case candidates generally show smaller or less consistent reductions than Placement Issue candidates. Consequently, Label Issue and Unresolved Case account for a larger share of the remaining review set even when their absolute counts do not increase. This compositional shift primarily reflects the sharper reduction in Placement Issue candidates rather than an increase in the absolute number of Label Issue or Unresolved Case candidates. The remaining review set therefore contains a greater proportion of Label Issue and Unresolved Case instances, for which the assigned class is not supported by the instance content.

Table~\ref{tab:idp-examples} illustrates the four diagnostic states using StackOverflow instances that share the same residual-mismatch condition: all 15 neighbors belong to class $B$, whereas the assigned class is $A$. Despite this identical neighborhood-level condition, the content judgments yield four distinct states. We use $\alpha=0.1$ only for illustration, not as an operating point for performance evaluation.

\begin{table}[t]
  \caption{\idp judgment examples from StackOverflow with Qwen3-Embedding-8B at $\alpha=0.1$. $J(x,A)$ and $J(x,B)$ denote the layout-independent judgments for the assigned class and the neighborhood-majority class, respectively. $\bullet$ indicates that the instance content supports the class, whereas $\circ$ indicates that it does not. In all four cases, all 15 neighbors are assigned class $B$, and none is assigned class $A$.}
  \label{tab:idp-examples}
  \centering
  \small
  \setlength{\tabcolsep}{2pt}
  \begin{tabular}{@{}llcc@{}}
    \toprule
    State & Instance & $J(x,A)$ & $J(x,B)$ \\
    \midrule
    \pir & \emph{Reading java .properties file from bash} & $\bullet$ & $\circ$ \\
    (placement) & $A$: \texttt{bash} \quad $B$: \texttt{spring} & & \\[2pt]
    \lir & \emph{Trying to get rsync to copy files back to svn} & $\circ$ & $\bullet$ \\
    (label) & $A$: \texttt{osx} \quad $B$: \texttt{svn} & & \\[2pt]
    \bcr & \emph{DB backup problem with Spring/Hibernate} & $\bullet$ & $\bullet$ \\
    (boundary) & $A$: \texttt{spring} \quad $B$: \texttt{hibernate} & & \\[2pt]
    \ucr & \emph{How come it is so expensive?} & $\circ$ & $\circ$ \\
    (unresolved) & $A$: \texttt{ajax} \quad $B$: \texttt{magento} & & \\
    \bottomrule
  \end{tabular}
\end{table}

The state-specific trajectories and examples clarify how neighborhood adjustment changes residual mismatches and why \idp remains necessary after adjustment. The comparatively large decrease in Placement Issue candidates is consistent with content-derived relational evidence reducing neighborhood configurations in which the neighborhood-majority class conflicts with the class supported by the instance content. By contrast, the presence of Label Issue, Boundary Case, and Unresolved Case instances in the remaining review set indicates that not every residual mismatch represents a neighborhood placement failure. Taken together, these results indicate that \kgn can narrow the evaluated gap between embedding-based neighborhoods and content-supported semantic relations, whereas \idp distinguishes the remaining mismatches into Placement Issue, Label Issue, Boundary Case, and Unresolved Case and shows why a residual mismatch should not automatically be interpreted as a projection failure.

\section{Conclusion}
\label{sec:conclusion}

% Embedding visualization faces a potential limitation: neighborhoods in the input embedding may not adequately reflect semantic relations in instance content. We propose \kgn, which uses triples shared between instance-specific KGs to reconstruct neighborhoods obtained from the input embedding, and introduce \idp, which interprets neighborhood--class mismatches remaining after adjustment based on instance content. Across all eight dataset--encoder combinations, \kgn improves both NMI and $\mathrm{LT}_{2D}$ while retaining 39.7--85.4\% of the original kNN neighbors. These improvements are observed across multiple KG similarity measures and with higher-capacity encoders, although some gains weaken under excessive guidance. Overall, the primary contribution of this work is \kgn, which converts relational knowledge derived from instance content into pairwise semantic evidence and complements neighborhood construction based on the input embedding without relying on predefined class labels. \idp supports this contribution by directly detecting neighborhood--class mismatches in each low-dimensional layout and diagnosing them according to whether the assigned class and the neighborhood-majority class adequately describe the instance content. This analysis shows that \kgn substantially reduces \pir candidates and thereby changes the composition of the residual review set.

We propose \kgn, which uses content-derived relational evidence to supplement embedding-based neighborhood construction, and introduce \idp to interpret residual mismatches at the instance level. Across the evaluated dataset--encoder conditions, the full guidance-strength sweep reveals condition-dependent increases in NMI and $\mathrm{LT}_{2D}$, with larger gains and more extensive neighborhood reconfiguration under the weak encoder conditions. These results indicate that \kgn can narrow the evaluated gap between embedding-based neighborhoods and content-supported semantic relations, while the variation across KG similarity measures, guidance strengths, and input embeddings shows that content-derived evidence complements rather than replaces the input embedding. The \idp analysis further shows that the largest reductions are concentrated in Placement Issue candidates, whereas Label Issue and Unresolved Case instances account for a larger share of the remaining review set. By distinguishing Placement Issue, Label Issue, Boundary Case, and Unresolved Case, \idp shows that a residual mismatch should not automatically be interpreted as a projection failure and instead identifies instances requiring state-specific interpretation.

% \section{Ethical Considerations}
% \label{sec:ethics}

\bibliographystyle{ACM-Reference-Format}
\bibliography{references}

\end{document}